\documentclass[journal]{IEEEtran}
\usepackage[cmex10]{amsmath}
\usepackage{amsmath}
\usepackage{mathtools}
\usepackage{amsfonts} 
\usepackage{siunitx}
\usepackage{algpseudocode}
\usepackage{algorithm}
\usepackage{booktabs}
\usepackage{threeparttable}
\usepackage{multirow}
\usepackage{makecell}
\usepackage{placeins}
\usepackage{mathrsfs,amsmath} 
\usepackage{fixltx2e}
\usepackage{color}
\usepackage{caption}
\usepackage[numbers,sort&compress]{natbib}
\usepackage{balance}

\newcommand{\G}{\mathcal{G}}
\newcommand{\V}{\mathcal{V}}
\newcommand{\E}{\mathcal{E}}
\newcommand{\W}{\mathcal{W}}
\newcommand{\N}{\mathcal{N}}
\newcommand{\K}{\mathcal{K}}
\newcommand{\Rbb}{\mathbb{R}}

\DeclareMathOperator{\prox}{prox}

\DeclareMathOperator{\sign}{sgn}

\DeclareSIUnit\Molar{\textsc{m}} 
\DeclareSIUnit{\pH}{pH}
\DeclareSIUnit{\pixel}{px}

\newcommand{\nauman}[1]{{\textcolor[rgb]{0,0,0}{#1}}}

\newcommand{\fn}[1]{{\textcolor[rgb]{0,0,0}{#1}}}

\title{
Adaptive Graph-based Total Variation for Tomographic Reconstructions
}

\author{Faisal Mahmood$^{*}$, Nauman Shahid$^{*}$, Ulf Skoglund$^{\dagger}$ and Pierre Vandergheynst$^{\dagger}$

\thanks{* Contributed Equally. $\dagger$ Equal Co-senior Authors.}
\thanks{F. Mahmood is with the Department of Biomedical Engineering, Johns Hopkins University (JHU), Baltimore, MD. U. Skoglund is with the Structural Cellular Biology Unit, Okinawa Institute of Science and Technology (OIST), 1919-1 Tancha, Onna, Okinawa 904-0495, Japan. N. Shahid and P. Vandergheynst are with the Signal Processing Laboratory 2 (LTS2), \'Ecole Polytechnique F\'ed\'erale de Lausanne (EPFL), STI IEL, Lausanne, CH-1015, Switzerland.}%
\thanks{F.M. and U.S. were supported by Japanese Government OIST Subsidy for Operations(Skoglund U.) under grant number 5020S7010020. F.M. was additionally supported by the OIST PhD Fellowship. N.S. and P.V. were supported by Swiss National Science Foundation (SNF) grant 200021\_154350/1.}
\thanks{Corresponding Author: F. Mahmood faisalm@jhu.edu}
\thanks{Images are best viewed in color in the electronic version of this manuscript.}
\thanks{This letter is accompanied with additional supplementary material.}

}

\begin{document}

\maketitle

\begin{abstract}
Sparsity exploiting image reconstruction (SER) methods have been extensively used with Total Variation (TV) regularization for tomographic reconstructions. Local TV methods fail to preserve texture details and often create additional artifacts due to over-smoothing. Non-Local TV (NLTV) methods have been proposed as a solution to this but they either lack continuous updates due to computational constraints or limit the locality to a small region.  In this paper we propose Adaptive Graph-based TV (AGTV). The proposed method goes beyond spatial similarity between different regions of an image being reconstructed by establishing a connection between similar regions in the entire image regardless of spatial distance. As compared to NLTV the proposed method is computationally efficient and involves updating the graph prior during every iteration making the connection between similar regions stronger. Moreover, it promotes sparsity in the wavelet and graph gradient domains. Since TV is a special case of graph TV the proposed method can also be seen as a generalization of SER and TV methods. 
\end{abstract}


\begin{IEEEkeywords}
Tomography, Total Variation, Graphs, Iterative Image Reconstruction, Non-local Total Variation
\end{IEEEkeywords}

\section{Introduction}
\IEEEPARstart{R}
ECONSTRUCTING tomographic densities from low-dose electron tomography (ET) or computed tomography (CT) data is an \textit{ill-posed inverse problem}. \fn{Low-dose is a constraint to prevent sample degradation in ET \cite{frank_electron_2008,leis_cryo-electron_2006}} and to reduce exposure to ionizing radiation in CT \cite{berrington_de_gonzalez_projected_2009,brenner2007computed,pearce2012radiation}. Such requirements are often met by collecting limited or low-contrast data which renders noisy and erroneous reconstructions. Iterative Image Reconstruction (IIR) methods \cite{fessler_statistical_2000,censor_finite_1983,qi2006iterative,skoglund_maximum-entropy_1996,rullgard_componentwise_2007} have proved to be more effective in handling noise when compared to analytical methods \cite{natterer_mathematics_1986,quinto2009electron,hsieh_computed_2009}. However, such methods are computationally inefficient. Initial IIR methods were algebraic in nature \cite{gordon_algebraic_1970,cimmino_calcolo_1938,hansen_air_2012,landweber_iteration_1951,brandt1986algebraic,strohmer2009randomized}. More recently sparsity exploiting reconstructions have been extensively used for image reconstruction. Such methods are often used with Total Variation (TV) regularization \cite{graff2015compressive,chen2008prior,song2007sparseness,ritschl2011improved,tang_performance_2009,tian2011low}. We refer to the joint \fn{Compressed Sensing} (CS) and TV setup as CSTV in the sequel. Recently, non-local TV (NLTV) \cite{lou2010image} has been shown to be much more efficient for \textit{inverse problems} \cite{peyre2008non,gilboa2008nonlocal,huang2012compressed,liu2016ticmr,jia20104d}. In contrast to simple TV, which takes into account the similarity of a region with only its neighboring regions, NLTV overcomes this limitation by associating a similarity measure of every region of an image with all other regions (full NLTV) \nauman{or a few regions in a spatial neighborhood (partial NLTV)}.

A primary short-coming of full NLTV is the high cost of associating a similarity measure between every pair of regions in an image ( $\mathcal{O}(n^4)$ for an $n \times n$ image). Hence, the similarity matrix constructed in the beginning from the initial estimate or \textit{prior} is not updated throughout the algorithm \cite{huang2012compressed,liu2016ticmr}.  \nauman{In order to overcome the computational complexity for adaptive updates, partial NLTV methods \cite{peyre2008non, lou2010image} tend to limit the nearest neighbors search to a local neighborhood of the pixel (hence we call them partial), which depends on a parameter $\delta$. For such methods, the computational cost drops down from $\mathcal{O}(n^4)$ to $\mathcal{O}(n^2 \delta^2)$, where $\delta \ll n$. However, it is quite probable that two spatially distant patches in an image are quite similar in structure. Thus, such methods lack the capability to model the pairwise relationships between the patches of an image on a global level.  The final reconstruction would be more faithful to the data if 1) the similarity matrix  is regularly updated during every iteration and 2) pairwise relationships are taken into account among all the patches of the image.}

\textbf{Introduction to Graphs}:
\nauman{Graphs, a discrete way of characterizing non-local variation methods, have emerged as a very powerful tool for signal modeling \cite{shuman_emerging_2012,sandryhaila2013discrete}.} A graph is represented as  a tuple $\mathcal{G}=\{ \V,\E,\mathcal{W}\}$, where $\V $ is a set of vertices, $\E$ a set of edges,
and $\W : \V \times \V \rightarrow \Rbb_+$ a weight function.    The weight matrix  $W$ is assumed to be non-negative, symmetric, and with a zero diagonal. Each entry  of the weight matrix $W \in \mathbb R^{|\V|\times |\V|}_+$ corresponds to the weight of the edge connecting the corresponding vertices: $W_{i,j} = \W (v_i,v_j)$ and if there is no edge between two vertices, the weight is set to $0$.    For a vertex $v_i\in \V$, the degree $d(i)$ is defined as the sum of the weights of incident edges: $d(i)=\sum_{j \leftrightarrow i } W_{i,j}$. 
 Let $D$ be the diagonal degree matrix with diagonal entries $D_{ii}=d(i)$, then the graph Laplacian $L$ is defined as  the difference of the weight matrix $W$ from the degree matrix $D$, thus $L = D - W$, which is referred to as combinatorial Laplacian. \fn{A more detailed account of the theory of signal processing on graphs can be found in seminal papers \cite{shuman_emerging_2012,sandryhaila2013discrete,perraudin2017stationary,teke2017extending,6879640}.}

\textbf{Contributions}: \nauman{Our previous work \cite{mahmood2016graph} has focused on using graph-based Total Variation for denoising the sinogram as a pre-processing step followed by using standard reconstruction methods such as SIRT or ART for reconstruction.} In this letter we propose Adaptive Graph Total Variation (AGTV) as a novel method for simultaneous reconstruction and denoising of tomographic data. Our \fn{proposed method} can be seen as a more sophisticated and adaptive form of full NLTV in the sense that it enjoys a relatively lower computational complexity by using an approximate $\K$-nearest neighbor search algorithm, where $\K$ is fixed. Due to a significant computational cost reduction, we can afford to update the graph in every iteration making the setup adaptive. Furthermore, our proposed method models the sparsity of the reconstructed image in: 1) Wavelet domain and 2) Graph gradient domain. These improvements lead to state-of-the art reconstruction results for both phantom data with known ground truth and real electron tomography data.   

\vspace{-1em}
\section{Adaptive Graph Total Variation (AGTV)}
Let $S \in \Re^{p \times q}$ be the sinogram corresponding to the projections of the sample $X \in \Re^{n \times n}$ being imaged,  where $p$ is the number of rays passing through $X$ and $q$ is the number of angular variations at which $X$ has been imaged.   Let $b \in \Re^{pq}$ be the vectorized measurements or projections $(b = vec(S))$, where $vec(\cdot)$ denotes the vectorization operation and $A \in \Re^{pq \times n^2}$ be the sparse projection operator. Then, the goal in a typical CT or ET based reconstruction method is to recover the vectorized sample $x = vec(X)$ from the projections $b$.  We propose:
 \vspace{-1.2em}

\begin{align}\label{eq:acsgt}
& \min_x \|Ax - b\|^2_2 + \lambda \|\Phi^* (x)\|_1 + \gamma \|\nabla_{\G} (x)\|_1,
\end{align}
where $\Phi $ is the  wavelet operator and $\Phi^*(x)$, where $*$ represents the adjoint operation, denotes the wavelet transform of $x$ and $ \|\nabla_{\G} (x)\|_1$  denotes the total variation of $x$ w.r.t graph $\G$.  The first two terms of the objective function above comprise the \textit{ sparse reconstruction} part of our method and model the sparsity of the wavelet coefficients. The second term, to which we refer as the \textit{graph total variation} (GTV) regularizer acts as an additional prior for denoising and smoothing. It can be expanded as:
 \vspace{-1.2em}

\begin{align*}\label{eq:gtv}
\nauman{ \|\nabla_{\G} (x)\|_1 = \sum_{i} \|\nabla_{\G } x_i\|_1 =  \sum_i {\sum_{j \in \N_i} \sqrt{W_{ij}}\|x_i - x_j\|_1}},
\end{align*} 
where the second sum runs over all the neighbors of $i$, denoted by $\N_i$. The above expression states that GTV involves the minimization of the sum of the gradients of the signals on the nodes of the graphs. In our case, we assume that the elements of the vector $x$ lie on the nodes of the graph $\G$ which are connected with the edges whose weights are $W_{ij}$. Thus, the minimization of the GTV would ensure that $x_i$ and $x_j$ possess similar values if $W_{ij}$ is high and dissimilar values if $W_{ij}$ is small or zero. As compared to standard TV, the structure of the sample $x$ is taken into account for reconstruction. It is  a well known fact that $l_1$ norm promotes sparsity, so the GTV can also be viewed as a regularization which promotes sparse graph gradients. This corresponds to enforcing a piecewise smoothness of the signal $x$ w.r.t graph $\G$. 
 
 The proposed method with GTV can be seen as a generalization of the compressed sensing and total variation based method studied in \cite{tang_performance_2009}. While, the standard TV minimizes the gradients of the signal $x$ w.r.t its spatial neighbors only, the GTV does so in a region which is not restricted only to the neighbors of the elements in $x$. Thus, the standard TV can be viewed as a specific case of the GTV, where the graph $\G_{grid}$ is a grid graph. In a grid graph $\G_{grid}$ of a sample $x$, the pixels are only connected to its spatial neighbors via unity weights. 

An important step for our method is to construct a graph $\G$ for GTV regularization.  Ideally, $\G$ should be representative of the reconstructed sample $x$, however, this is unknown before the reconstruction. To cater this problem, we propose to construct $\G$ from the patches of an initial naive estimate of the sample $x_{fbp}$ using analytical filtered back projection (FBP). 
 In the first step $x_{fbp} \in \Rbb^{n \times n}$ is divided into $n^2$ overlapping patches. Let $s_i$ be the patch of size $l \times l$ centered at the $i^{th}$ pixel of $x_{fbp}$ and assume that all patches are vectorized, \textit{i.e}, $s_i \in \Re^{l^2}$. In the second step the search for the closest neighbors for all vectorized patches is performed using the Euclidean distance metric. \nauman{For two patches $s_i,s_j$, the distance metric is defined as $\|s_i - s_j\|_2$}. Each  $s_i$ is connected to its $\K$ nearest neighbors $s_j$ \nauman{only}, resulting in $|\mathcal{E}|$ number of connections. \nauman{This is realized by computing all the pairwise distances between all possible patches $s_i, s_j$ and then keeping only the most relevant $\K$ neighbors.} In the third step the graph weight matrix $W$ is computed using the Gaussian kernel weighting scheme, for which the parameter $\sigma$ is set experimentally as the average distance of the connected samples. \nauman{Hence, for the patches $s_i, s_j$, the weighting scheme is defined as $W_{i,j} = \exp(-\|s_i - s_j\|^2_2 / \sigma^2)$. Finally, the combinatorial Laplacian is computed.}
 
  Note that the computation of the weight matrix $W$ for graph $\G$ costs $\mathcal{O}(n^4)$. \nauman{ As mentioned earlier,  our goal is to aovid this cost and update the graph in every iteration. For this purpose, we propose to make the graph construction efficient by using an approximate nearest neighbor search algorithm by using the FLANN library (Fast Library for Approximate Nearest Neighbors searches in high dimensional spaces) \cite{muja2014scalable}. This reduces the cost of graph construction from $\mathcal{O}(n^4)$ to $\mathcal{O}(n^2 \log(n))$.}

\nauman{The above description refers only to the non-adaptive part, where the graph $\G$ is fixed. It is important to point out that the initial estimate of the graph $\G$, obtained via the filtered back projection $x_{fbp}$ is not very faithful to the final solution $x$. As $x$ is being refined in every iteration, it is natural to update the graph $\G$ as well in every iteration. This simultaneous update of the graph $\G$ corresponds to the adaptive part of the proposed algorithm and its significance has been explained in detail in the supplement with this letter.} 
 \begin{figure}
\centering
\includegraphics[width=3.5in]{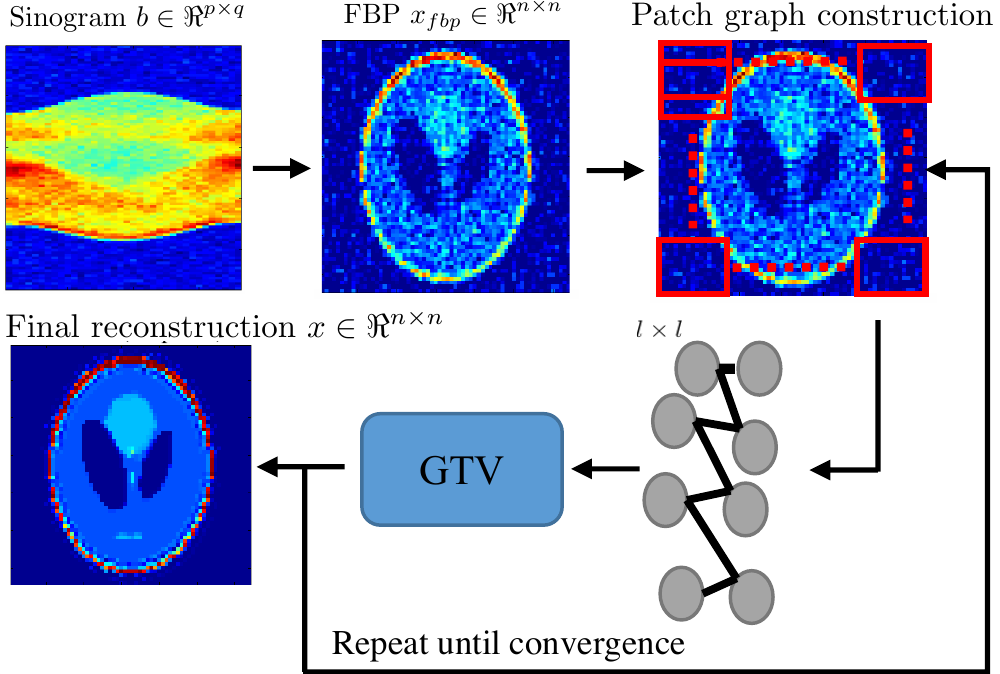}
\caption{The complete methodology for AGTV. The input projections $b \in \Rbb^{p\times q}$ is first used to obtained a filtered back projection (FBP) $x_{fbp} \in \Rbb^{n \times n}$. It is then used to construct the initial patch graph $\G$ to be used by the GTV method. The output of GTV is used to refine / reconstruct the graph and this process is repeated until convergence.}
\label{fig:flow} 
\vspace{-1.5em}
\end{figure}

 \vspace{-0.8em}
\section{Optimization Solution}\label{sec:opt}
In the spirit of similar non-graph methods such as \cite{tang_performance_2009}, we refer to eq. \eqref{eq:acsgt} without the graph update as Compressed Sensing and Graph Total Variation or simply GTV. We make use of forward backward based primal dual method \cite{komodakis2015playing, combettes_proximal_2011} to solve GTV and then update the graph from the obtained sample in every iteration, until convergence. The complete algorithm with graph updates is called Adaptive Graph Total Variation (AGTV). The main steps of this algorithm are visualized in Fig. \ref{fig:flow}.

\nauman{The first term of Eq. 1, $f:\mathbb{R}^{n^2}\rightarrow \mathbb{R}$ is a convex differentiable  function defined as $f(x) = \|Ax-b\|^2_2 $.  This function has a $\beta$-Lipschitz continuous gradient $\nabla_f(x) = 2 A^{\top}(Ax-b).$ Note that $\beta = 2  \| A\|_2$ where $\|A \|_2$ is the  spectral norm (or maximum eigenvalue) of $A$. The constant $\beta$ has important implications in deciding the time step in iterative optimization methods. Let $\tau_1,\tau_2,\tau_3$ be the step size parameters.} \nauman{As a rule of thumb,  these parameters are typically set to the inverse of the Lipschitz constant $\beta$. Hence, we set $\tau_1,\tau_2, \tau_3$ proportional to $1/\beta$. Furthermore, note that these parameters are independent of the regularization parameters $\lambda$ and $\gamma$}.

\nauman{The proximal operator of the second function $g =  \lambda \|\Phi^* (x)\|_1$ (in Eq. \ref{eq:acsgt}) is the $\ell_1$ soft-thresholding of the wavelet coefficients given by the elementwise operations}.
\begin{equation}
\nauman{\prox_{\tau_1 g }(x) = \sign(x) \circ \max (|x|-\tau_1\lambda ,0),}
\end{equation}

\nauman{The third term in Eq. \eqref{eq:acsgt} $h: \mathbb{R}^{|\mathcal{E}|n}\rightarrow \mathbb{R}$, where $|\mathcal{E}|$ denotes the cardinality of $\mathcal{E}$ the set of edges in $\G$, is a convex function defined as $h(D) = \gamma \|D\|_1 $. The proximal operator, where $\circ$ denotes the Hadamard product and $D = \nabla_{\G } x$. is:}
\begin{equation}
\nauman{\prox_{\tau_2 h }(D) = \sign(D) \circ \max (|D|-\tau_2 \gamma ,0),}
\end{equation}
Using these tools, we can use the forward backward based primal dual approach presented in \cite{komodakis2015playing},  for AGTV, to define Algorithm \ref{CHalgorithm}  where $\epsilon$ the stopping tolerance, \fn{$I,J$ define the maximum number of iterations} and $\delta$ is a very small number to avoid a possible division by $0$.  

\begin{algorithm}
\caption{Forward-backward primal dual for AGTV}
\label{CHalgorithm}
\begin{algorithmic}
\State $x_0 = x_{fbp}$
\State 1. INPUT: $U_0 = x_{0}$, $V_0 = \nabla_{\mathcal{G}}x_{0}$, $\epsilon > 0$
\For{\nauman{$i = 0, \dots I - 1$}}
\For{ \nauman{$j = 0,\dots J-1$ }}
\State a. $P_{j} = \Phi(\prox_{\tau_1  g} \left(\Phi^*(U_{j}) - \tau_1 \Phi^*\left(\nabla_f(U_{j}) + \nabla_{\mathcal{G}}^* V_j \right) \right)$
\State b. $T_j = V_j + \tau_2 \nabla_{\mathcal{G}}(2P_j-U_j)$
\State c. $Q_{j} = T_j - \tau_2 \prox_{\frac{1}{\tau_2} h} \left( \frac{1}{\tau_2} T_j \right) $
\State d. $(U_{j+1},V_{j+1}) = (U_{j},V_{j}) + \tau_3   ((P_{j},Q_{j}) -  (U_{j},V_{j})) $
\If{$\frac{\|U_{j+1} - U_{j}\|_F^2}{\| U_{j}\|_F^2+\delta}<\epsilon$ and $\frac{\|V_{j+1} - V_{j}\|_F^2}{\| V_{j}\|_F^2+\delta}<\epsilon$}
\State BREAK\textbf{}
\EndIf
\EndFor
 \State 2. $x_i = U_{j+1}$
 \State 3. Construct patch graph $\G$ from $x_i$
 \If{\nauman{$\frac{\|x_{i} - x_{i-1}\|_F^2}{\| x_{i}\|_F^2+\delta}<\epsilon$}}
 \State \nauman{BREAK }
 \EndIf
 \EndFor
\State OUTPUT: $x_i$
\end{algorithmic}
\end{algorithm}
\textbf{Complexity}: We use the Fast Approximate Neartest Neighbors search algorithm (FLANN) \cite{muja2014scalable}, whose  computational complexity for $n^2$  patches of size $l^2$ each and fixed $K$ is $\mathcal{O}(n^2\log(n))$. Let $J$ and $I$ denote the maximum number of iterations for the algorithm to converge, then the computational cost of our algorithm  is $\mathcal{O}(J |\mathcal{E}| I)$, where $|\mathcal{E}|$ denotes the number of non-zeros edges in the graph $\G$. For a $\K$-nearest neighbors graph $|\mathcal{E}| \approx \K n^2$ so the computational complexity of our algorithm is  linear in the size of the data sample $n^2$, \textit{i.e}  $\mathcal{O}(J \K n^2 I)$. The graph $\G$ needs to be updated once in every outer iteration of the algorithm $I$, thus the overall complexity is $\mathcal{O}(IJ \K n^2 + I n^2\log(n))$. \fn{Dropping constants GTV scales with $\mathcal{O}(n^2)$ and AGTV scales with $\mathcal{O}(n^2(1 + \log(n))$.}


\vspace{-0em}
\section{Experimental Results}
\vspace{-0.5em}
\begin{figure*}
\centering
\includegraphics[width=0.95\textwidth,height=13.5cm,keepaspectratio]{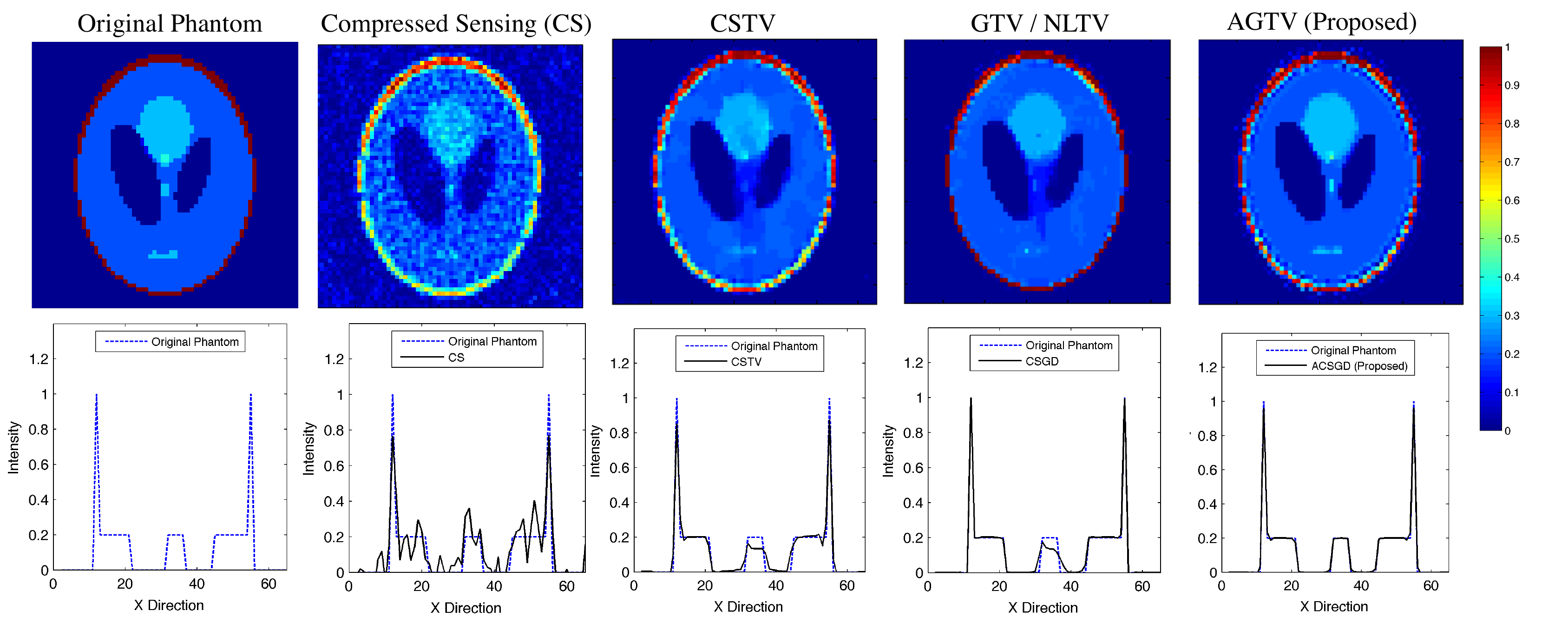}
\caption{Comparative analysis of reconstructing Shepp-Logan using various reconstruction methods. The sinogram of a $64\times64$ Shepp-Logan phantom corrupted with 10\% Poission noise was reconstructed using FBP (Linearly interpolated, Cropped Ram-Lak filter); CSTV ($\lambda=0.5$, $\gamma=0.1$, Prior: FBP, Stopping Criteria = 100 iterations); GTV/NLTV ($\lambda=0.5$, $\gamma=0.2$, Prior: Patch Graph from FBP, Stopping Criteria = 100 iterations); AGTV ($\lambda=0.5$, $\gamma=1$, Prior: Patch Graph from FBP updated every iteration, $I$ and $J$ in Algorithm 1 set to 30). AGTV gives a better intensity profile as compared to all other methods while preserving the edges. }
\label{fig:reconstructions}
\vspace{-0.5em}
\end{figure*}

\begin{figure*}
\centering
\includegraphics[width=\textwidth, height = 0.35\textwidth]{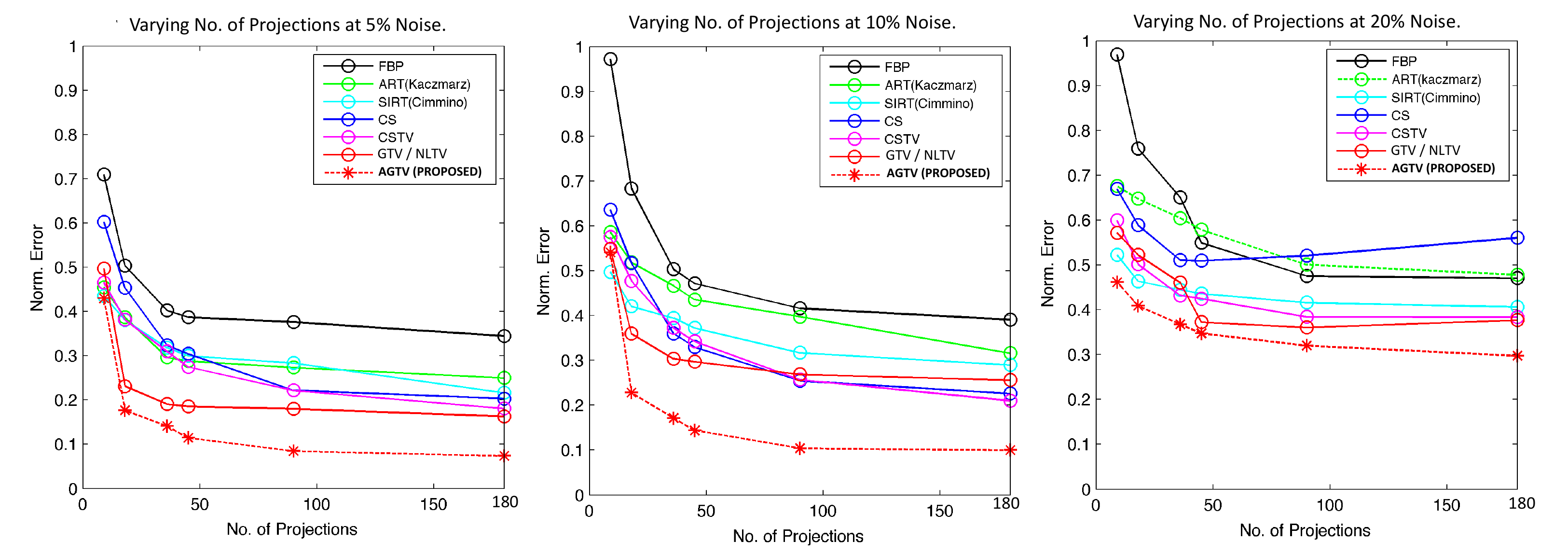}
\caption{Comparative analysis of reconstructing a Shepp-Logan phantom using various reconstruction methods at 5\% and 10\% Poisson noise. FBP (Linearly interpolated, Cropped Ram-Lak filter); ART (Kaczmarz/Randomized Kaczmarz, Relaxation Parameter $(\eta)$ = 0.25, Prior: FBP, Stopping Criteria = 100 iterations); SIRT (Cimmino/SART, $(\eta)$ = 0.25, Prior: FBP, Stopping Criteria = 100 iterations); CS (500 Iterations, Prior: FBP); CSTV ($\lambda=0.5$, $\gamma=0.1$, Prior: FBP, Stopping Criteria = 100 iterations); GTV ($\lambda=0.5$, $\gamma=0.2$, Prior: Patch Graph from FBP, Stopping Criteria = 100 iterations); AGTV ($\lambda=0.5$, $\gamma=1$, Prior: Patch Graph from FBP updated every iteration, $I$ and $J$ in Algorithm 1 set to 30). }
\label{fig:phantoms}
\vspace{-1.5em}
\end{figure*}

To test the performance of our AGTV method, we perform reconstructions for many different types of phantoms from different number of projections with varying levels of Poisson noise, using  GSPBox \cite{perraudin2014gspbox},  UNLocBox  \cite{perraudin2014unlocbox} and AIRTools \cite{hansen_air_2012}. Reconstructions were judged on a $\ell_2$ reconstruction error metric. We compare the performance of AGTV with many state-of-the-art iterative and convex optimization based algorithms, which include  FBP, ART (Kaczmarz), SIRT (Cimmino), CS, CSTV and GTV (FLANN approximation of NLTV). \nauman{All hyperparameters were tuned for best performance.}

Each of these methods has its own model parameters, which need to be set or tuned in an appropriate manner. ART (Kaczmarz) and SIRT (Cimmino) were performed using FBP as \textit{a priori}. \nauman{The stopping criteria for ART and SIRT was set to 100 iterations and the relaxation parameter $(\eta)$ was tuned to achieve the best result.} For the graph based reconstruction (GTV, AGTV) a graph prior $\G$ was generated by dividing the result from FBP into patches as explained previously. For example, for a Shepp-Logan phantom of size $64 \times 64$, the graph was constructed by dividing it into $64 \times 64 = 4096$ overlapping patches of size $3\times 3$, $\K = 15$ and setting $\sigma$ for the weight matrix to the average distance of the 15-nearest neighbors.  For Algorithm 1, we set $I = J = 30$ and the convergence parameters $\tau_1, \tau_2, \tau_3$ were set automatically by UNLocBox.  It is worth mentioning here that GTV is a faster method of implementing NLTV by using $
\K$-nearest neighbors graph approximation. Thus the GTV and NLTV based regularization are approximately equivalent in performance. Therefore, we did not include comparisons with NLTV.

To explain the performance of our model in detail  we reconstructed a $64 \times 64$ Shepp-Logan \cite{shepp_fourier_1974} phantom from 36 erroneous projections. A sinogram $S$ was built by projecting the phantom using Radon transform and 36 equally spaced projections were collected from 0 to 180 degrees. The sinogram was then corrupted with 10\% Poission noise. Fig. \ref{fig:phantoms} provides a detailed comparison of the reconstruction of Shepp-Logan phantom via various algorithms along with the intensity profiles plotted underneath each of the reconstructions.  It can be seen  that AGTV performs better than GTV and CSTV. A similar experimental setup was repeated by reconstructing a $128\times128$ Torso phantom from 36 erroneous projections corrupted with 5\% Gaussian normalized noise and similar results were achieved (Fig. 3 in the Supplement).  A graphical comparison for the reconstruction of Shepp-Logan using various reconstruction methods at varying number of projections and noise levels has been given in Fig. \ref{fig:phantoms}. \fn{AGTV shows promising results even with limited data reconstructions and outperforms many other state-of-the-art reconstruction and denoising methods.} A more detailed analysis of these results has been presented in the supplement.

\vspace{-0.5em}
\section{Conclusions}
\vspace{-0.5em}
Similar to NLTV our proposed method goes beyond spatial similarity between different regions of an image being reconstructed by establishing a connection between similar regions in the image regardless of spatial distance. However, it is much more scalable and computationally efficient because it uses the approximate nearest neighbor search algorithm for graph construction, making it more likely to be adapted in a clinical setting. \nauman{Beyond NLTV, our proposed approach is adaptive. The non-local graph prior is updated every iteration making the connection between similar regions stronger, thus, improving the overall reconstruction quality. Since TV is a special case of graph TV the proposed method can be seen as a generalization of CS and TV methods. Shortcomings of the proposed method include decreased graph quality due to approximations and tedious hyperparameter tuning. }

\ifCLASSOPTIONcaptionsoff
  \newpage
\fi

\balance

\bibliographystyle{IEEEtran}
\bibliography{bibfile.bib}

\begin{thebibliography}{10}
\def\url#1{}
\csname url@samestyle\endcsname
\providecommand{\newblock}{\relax}
\providecommand{\bibinfo}[2]{#2}
\providecommand{\BIBentrySTDinterwordspacing}{\spaceskip=0pt\relax}
\providecommand{\BIBentryALTinterwordstretchfactor}{4}
\providecommand{\BIBentryALTinterwordspacing}{\spaceskip=\fontdimen2\font plus
\BIBentryALTinterwordstretchfactor\fontdimen3\font minus
  \fontdimen4\font\relax}
\providecommand{\BIBforeignlanguage}[2]{{%
\expandafter\ifx\csname l@#1\endcsname\relax
\typeout{** WARNING: IEEEtran.bst: No hyphenation pattern has been}%
\typeout{** loaded for the language `#1'. Using the pattern for}%
\typeout{** the default language instead.}%
\else
\language=\csname l@#1\endcsname
\fi
#2}}
\providecommand{\BIBdecl}{\relax}
\BIBdecl

\bibitem{frank_electron_2008}
J.~Frank, \emph{Electron tomography: methods for three-dimensional
  visualization of structures in the cell}.\hskip 1em plus 0.5em minus
  0.4em\relax Springer Science \& Business Media, 2008.

\bibitem{leis_cryo-electron_2006}
\BIBentryALTinterwordspacing
A.~Leis, M.~Beck, M.~Gruska, C.~Best, R.~Hegerl, and J.~Leis, ``Cryo-electron
  tomography of biological specimens,'' \emph{IEEE Signal Processing Magazine},
  vol.~23, no.~3, pp. 95--103, May 2006.
  \url{http://ieeexplore.ieee.org/lpdocs/epic03/wrapper.htm?arnumber=1628882}
\BIBentrySTDinterwordspacing

\bibitem{berrington_de_gonzalez_projected_2009}
\BIBentryALTinterwordspacing
A.~Berrington~de Gonz{\'a}lez, ``\BIBforeignlanguage{en}{Projected {Cancer}
  {Risks} {From} {Computed} {Tomographic} {Scans} {Performed} in the {United}
  {States} in 2007},'' \emph{\BIBforeignlanguage{en}{Archives of Internal
  Medicine}}, vol. 169, no.~22, p. 2071, Dec. 2009.
  \url{http://archinte.jamanetwork.com/article.aspx?doi=10.1001/archinternmed.2009.440}
\BIBentrySTDinterwordspacing

\bibitem{brenner2007computed}
D.~J. Brenner and E.~J. Hall, ``Computed tomography---an increasing source of
  radiation exposure,'' \emph{N Engl J Med}, vol. 357, pp. 2277--84, 2007.

\bibitem{pearce2012radiation}
M.~S. Pearce, J.~A. Salotti, M.~P. Little, K.~McHugh, C.~Lee, K.~P. Kim, N.~L.
  Howe, C.~M. Ronckers, P.~Rajaraman, A.~W. Craft \emph{et~al.}, ``Radiation
  exposure from ct scans in childhood and subsequent risk of leukaemia and
  brain tumours: a retrospective cohort study,'' \emph{The Lancet}, vol. 380,
  no. 9840, pp. 499--505, 2012.

\bibitem{fessler_statistical_2000}
J.~A. Fessler, ``Statistical image reconstruction methods for transmission
  tomography,'' \emph{Handbook of medical imaging}, vol.~2, pp. 1--70, 2000.

\bibitem{censor_finite_1983}
\BIBentryALTinterwordspacing
Y.~Censor, ``Finite series-expansion reconstruction methods,''
  \emph{Proceedings of the IEEE}, vol.~71, no.~3, pp. 409--419, 1983.
  \url{http://ieeexplore.ieee.org/lpdocs/epic03/wrapper.htm?arnumber=1456866}
\BIBentrySTDinterwordspacing

\bibitem{qi2006iterative}
J.~Qi and R.~M. Leahy, ``Iterative reconstruction techniques in emission
  computed tomography,'' \emph{Physics in medicine and biology}, vol.~51,
  no.~15, p. R541, 2006.

\bibitem{skoglund_maximum-entropy_1996}
\BIBentryALTinterwordspacing
U.~Skoglund, L.-G. {\"O}fverstedt, R.~M. Burnett, and G.~Bricogne,
  ``\BIBforeignlanguage{en}{Maximum-{Entropy} {Three}-{Dimensional}
  {Reconstruction} with {Deconvolution} of the {Contrast} {Transfer}
  {Function}: {A} {Test} {Application} with {Adenovirus}},''
  \emph{\BIBforeignlanguage{en}{Journal of Structural Biology}}, vol. 117,
  no.~3, pp. 173--188, Nov. 1996.
  \url{http://linkinghub.elsevier.com/retrieve/pii/S1047847796900817}
\BIBentrySTDinterwordspacing

\bibitem{rullgard_componentwise_2007}
\BIBentryALTinterwordspacing
H.~Rullg{\aa}rd, O.~{\"O}ktem, and U.~Skoglund, ``A componentwise iterated
  relative entropy regularization method with updated prior and regularization
  parameter,'' \emph{Inverse Problems}, vol.~23, no.~5, pp. 2121--2139, Oct.
  2007.
  \url{http://stacks.iop.org/0266-5611/23/i=5/a=018?key=crossref.0fc32a8fc4c7da490a96eae7f8bf45ba}
\BIBentrySTDinterwordspacing

\bibitem{natterer_mathematics_1986}
F.~Natterer, \emph{The mathematics of computerized tomography}.\hskip 1em plus
  0.5em minus 0.4em\relax Siam, 1986, vol.~32.

\bibitem{quinto2009electron}
E.~T. Quinto, U.~Skoglund, and O.~{\"O}ktem, ``Electron lambda-tomography,''
  \emph{Proceedings of the National Academy of Sciences}, vol. 106, no.~51, pp.
  21\,842--21\,847, 2009.

\bibitem{hsieh_computed_2009}
J.~Hsieh, ``Computed tomography: principles, design, artifacts, and recent
  advances.''\hskip 1em plus 0.5em minus 0.4em\relax SPIE Bellingham, WA, 2009.

\bibitem{gordon_algebraic_1970}
\BIBentryALTinterwordspacing
R.~Gordon, R.~Bender, and G.~T. Herman, ``\BIBforeignlanguage{en}{Algebraic
  {Reconstruction} {Techniques} ({ART}) for three-dimensional electron
  microscopy and {X}-ray photography},'' \emph{\BIBforeignlanguage{en}{Journal
  of Theoretical Biology}}, vol.~29, no.~3, pp. 471--481, Dec. 1970.
  \url{http://linkinghub.elsevier.com/retrieve/pii/0022519370901098}
\BIBentrySTDinterwordspacing

\bibitem{cimmino_calcolo_1938}
G.~Cimmino and C.~N. delle Ricerche, \emph{Calcolo approssimato per le
  soluzioni dei sistemi di equazioni lineari}.\hskip 1em plus 0.5em minus
  0.4em\relax Istituto per le applicazioni del calcolo, 1938.

\bibitem{hansen_air_2012}
P.~C. Hansen and M.~Saxild-Hansen, ``{AIR} tools---a {MATLAB} package of
  algebraic iterative reconstruction methods,'' \emph{Journal of Computational
  and Applied Mathematics}, vol. 236, no.~8, pp. 2167--2178, 2012.

\bibitem{landweber_iteration_1951}
L.~Landweber, ``An iteration formula for {Fredholm} integral equations of the
  first kind,'' \emph{American journal of mathematics}, vol.~73, no.~3, pp.
  615--624, 1951.

\bibitem{brandt1986algebraic}
A.~Brandt, ``Algebraic multigrid theory: The symmetric case,'' \emph{Applied
  mathematics and computation}, vol.~19, no.~1, pp. 23--56, 1986.

\bibitem{strohmer2009randomized}
T.~Strohmer and R.~Vershynin, ``A randomized kaczmarz algorithm with
  exponential convergence,'' \emph{Journal of Fourier Analysis and
  Applications}, vol.~15, no.~2, pp. 262--278, 2009.

\bibitem{graff2015compressive}
C.~G. Graff and E.~Y. Sidky, ``Compressive sensing in medical imaging,''
  \emph{Applied optics}, vol.~54, no.~8, pp. C23--C44, 2015.

\bibitem{chen2008prior}
G.-H. Chen, J.~Tang, and S.~Leng, ``Prior image constrained compressed sensing
  (piccs): a method to accurately reconstruct dynamic ct images from highly
  undersampled projection data sets,'' \emph{Medical physics}, vol.~35, no.~2,
  pp. 660--663, 2008.

\bibitem{song2007sparseness}
J.~Song, Q.~H. Liu, G.~A. Johnson, and C.~T. Badea, ``Sparseness prior based
  iterative image reconstruction for retrospectively gated cardiac micro-ct,''
  \emph{Medical physics}, vol.~34, no.~11, pp. 4476--4483, 2007.

\bibitem{ritschl2011improved}
L.~Ritschl, F.~Bergner, C.~Fleischmann, and M.~Kachelrie{\ss}, ``Improved total
  variation-based ct image reconstruction applied to clinical data,''
  \emph{Physics in medicine and biology}, vol.~56, no.~6, p. 1545, 2011.

\bibitem{tang_performance_2009}
\BIBentryALTinterwordspacing
J.~Tang, B.~E. Nett, and G.-H. Chen, ``Performance comparison between total
  variation ({TV})-based compressed sensing and statistical iterative
  reconstruction algorithms,'' \emph{Physics in Medicine and Biology}, vol.~54,
  no.~19, pp. 5781--5804, Oct. 2009.
  \url{http://stacks.iop.org/0031-9155/54/i=19/a=008?key=crossref.e211e5c298c689f9b653c24e6751353e}
\BIBentrySTDinterwordspacing

\bibitem{tian2011low}
Z.~Tian, X.~Jia, K.~Yuan, T.~Pan, and S.~B. Jiang, ``Low-dose ct reconstruction
  via edge-preserving total variation regularization,'' \emph{Physics in
  medicine and biology}, vol.~56, no.~18, p. 5949, 2011.

\bibitem{lou2010image}
Y.~Lou, X.~Zhang, S.~Osher, and A.~Bertozzi, ``Image recovery via nonlocal
  operators,'' \emph{Journal of Scientific Computing}, vol.~42, no.~2, pp.
  185--197, 2010.

\bibitem{peyre2008non}
G.~Peyr{\'e}, S.~Bougleux, and L.~Cohen, ``Non-local regularization of inverse
  problems,'' in \emph{European Conference on Computer Vision}.\hskip 1em plus
  0.5em minus 0.4em\relax Springer, 2008, pp. 57--68.

\bibitem{gilboa2008nonlocal}
G.~Gilboa and S.~Osher, ``Nonlocal operators with applications to image
  processing,'' \emph{Multiscale Modeling \&amp; Simulation}, vol.~7, no.~3,
  pp. 1005--1028, 2008.

\bibitem{huang2012compressed}
J.~Huang and F.~Yang, ``Compressed magnetic resonance imaging based on wavelet
  sparsity and nonlocal total variation,'' in \emph{2012 9th IEEE International
  Symposium on Biomedical Imaging (ISBI)}.\hskip 1em plus 0.5em minus
  0.4em\relax IEEE, 2012, pp. 968--971.

\bibitem{liu2016ticmr}
J.~Liu, H.~Ding, S.~Molloi, X.~Zhang, and H.~Gao, ``Ticmr: Total image
  constrained material reconstruction via nonlocal total variation
  regularization for spectral ct,'' 2016.

\bibitem{jia20104d}
X.~Jia, Y.~Lou, B.~Dong, Z.~Tian, and S.~Jiang, ``4d computed tomography
  reconstruction from few-projection data via temporal non-local
  regularization,'' in \emph{International Conference on Medical Image
  Computing and Computer-Assisted Intervention}.\hskip 1em plus 0.5em minus
  0.4em\relax Springer, 2010, pp. 143--150.

\bibitem{shuman_emerging_2012}
D.~I. Shuman, S.~K. Narang, P.~Frossard, A.~Ortega, and P.~Vandergheynst, ``The
  {Emerging} {Field} of {Signal} {Processing} on {Graphs}: {Extending}
  {High}-{Dimensional} {Data} {Analysis} to {Networks} and {Other} {Irregular}
  {Domains},'' \emph{arXiv preprint arXiv:1211.0053}, 2012.

\bibitem{sandryhaila2013discrete}
A.~Sandryhaila and J.~M. Moura, ``Discrete signal processing on graphs,''
  \emph{IEEE transactions on signal processing}, vol.~61, no.~7, pp.
  1644--1656, 2013.

\bibitem{perraudin2017stationary}
N.~Perraudin and P.~Vandergheynst, ``Stationary signal processing on graphs,''
  \emph{IEEE Transactions on Signal Processing}, vol.~65, no.~13, pp.
  3462--3477, 2017.

\bibitem{teke2017extending}
O.~Teke and P.~Vaidyanathan, ``Extending classical multirate signal processing
  theory to graphs—part i: Fundamentals,'' \emph{IEEE Transactions on Signal
  Processing}, vol.~65, no.~2, pp. 409--422, 2017.

\bibitem{6879640}
A.~Sandryhaila and J.~M.~F. Moura, ``Big data analysis with signal processing
  on graphs: Representation and processing of massive data sets with irregular
  structure,'' \emph{IEEE Signal Processing Magazine}, vol.~31, no.~5, pp.
  80--90, Sept 2014.

\bibitem{mahmood2016graph}
F.~Mahmood, N.~Shahid, P.~Vandergheynst, and U.~Skoglund, ``Graph-based
  sinogram denoising for tomographic reconstructions,'' in \emph{Engineering in
  Medicine and Biology Society (EMBC), 2016 IEEE 38th Annual International
  Conference of the}.\hskip 1em plus 0.5em minus 0.4em\relax IEEE, 2016, pp.
  3961--3664.

\bibitem{muja2014scalable}
M.~Muja and D.~G. Lowe, ``Scalable nearest neighbor algorithms for high
  dimensional data,'' \emph{IEEE Transactions on Pattern Analysis and Machine
  Intelligence}, vol.~36, no.~11, pp. 2227--2240, 2014.

\bibitem{komodakis2015playing}
N.~Komodakis and J.-C. Pesquet, ``Playing with duality: An overview of recent
  primal? dual approaches for solving large-scale optimization problems,''
  \emph{IEEE Signal Processing Magazine}, vol.~32, no.~6, pp. 31--54, 2015.

\bibitem{combettes_proximal_2011}
P.~L. Combettes and J.-C. Pesquet, ``Proximal splitting methods in signal
  processing,'' in \emph{Fixed-point algorithms for inverse problems in science
  and engineering}.\hskip 1em plus 0.5em minus 0.4em\relax Springer, 2011, pp.
  185--212.

\bibitem{perraudin2014gspbox}
N.~Perraudin, J.~Paratte, D.~Shuman, V.~Kalofolias, P.~Vandergheynst, and D.~K.
  Hammond, ``Gspbox: A toolbox for signal processing on graphs,'' \emph{arXiv
  preprint arXiv:1408.5781}, 2014.

\bibitem{perraudin2014unlocbox}
N.~Perraudin, D.~Shuman, G.~Puy, and P.~Vandergheynst, ``Unlocbox a matlab
  convex optimization toolbox using proximal splitting methods,'' \emph{arXiv
  preprint arXiv:1402.0779}, 2014.

\bibitem{shepp_fourier_1974}
\BIBentryALTinterwordspacing
L.~A. Shepp and B.~F. Logan, ``The {Fourier} reconstruction of a head
  section,'' \emph{IEEE Transactions on Nuclear Science}, vol.~21, no.~3, pp.
  21--43, Jun. 1974.
  \url{http://ieeexplore.ieee.org/lpdocs/epic03/wrapper.htm?arnumber=6499235}
\BIBentrySTDinterwordspacing

\end{thebibliography}

\onecolumn
\clearpage
\newpage

\vspace{-5em}
\appendix
\section*{A. Working Explanation of AGTV}\label{sec:explain}

\begin{figure}[h!]
\centering
\includegraphics[width=6in]{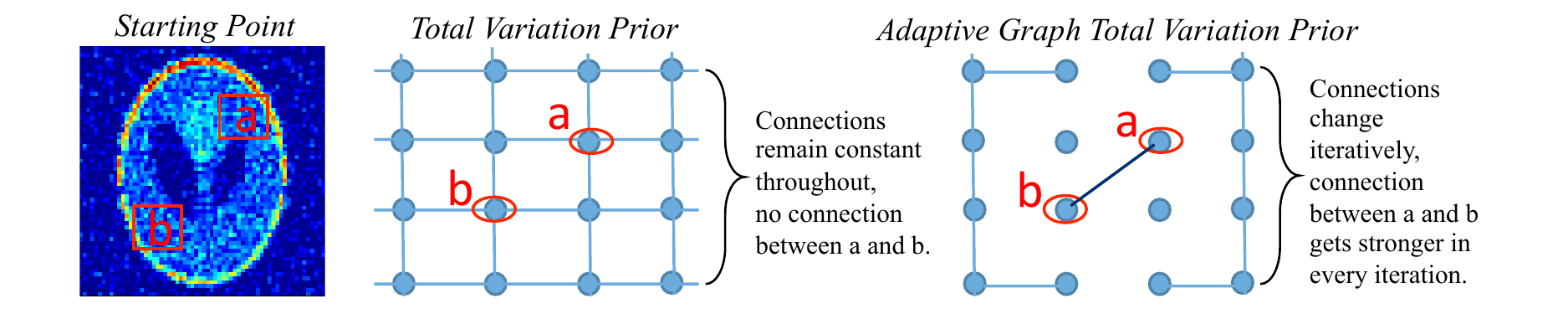}
\caption{A comparison of the Total Variation (TV) and Adaptive Graph Total Variation (AGTV) priors for the methods CSGT and AGTV. The TV prior does not connect patches `a' and `b' which possess structural similarity, whereas the GTV prior connects them because the $\K$-nearest neighbor graph is not restricted to spatial neighbors only. Furthermore, this connection keeps getting stronger due to iterative removal of noise and graph updates in every iteration.}
\label{fig:demo} 
\vspace{-1.2em}
\end{figure}

We present a simple example to motivate the use of AGTV rather than simple GTV and CSTV. Clearly, the compressed sensing part of all these methods is responsible for retrieving the sample $x$ from the projections $b$. Thus, our comparison study is focused on the two regularizers, \textit{i.e}, Adaptive Graph Total Variation (AGTV) and Total Variation (TV). Consider the example of a Shepp-Logan Phantom as shown in top leftmost plot of Fig. \ref{fig:reconstructions}. The goal is to recover this phantom from its noisy projections so that the recovered sample is faithful to its original clean version. \nauman{The CSTV method requires a TV prior to recover the sample while the GTV method requires a graph total variation prior for the recovery.} Both methods need an initial estimate for the construction of this prior, therefore, for the ease of demonstration we use the filtered back projection (FBP) as an initial estimate of the sample. Recall that our proposed method decomposes the FBP into $n \times n$ patches of size $l \times l$ each. Let $(i,j)$ denote the (horizontal, vertical) position of the center of each patch then: 1) For the total variation, each patch $s_{i,j}$ is connected to its spatial neighbors only, \textit{i.e}, $s_{i+1,j}, s_{i-1,j}, s_{i,j+1}, s_{i,j-1}$, as shown in Fig. \ref{fig:demo}. These connections are fixed throughout the algorithm. 2) For the graph total variation, each patch $s_{i,j}$ is only connected to the patches which are among the $\K$ nearest neighbors. Note that unlike TV the connected patches can be spatially far from each other. Now let us take the example of two patches `a' and `b' as labeled in the FBP of Fig. \ref{fig:demo}. Comparing with the clean phantom in Fig. \ref{fig:reconstructions} it is obvious that these patches should possess the same texture at the end of the reconstruction algorithm. Therefore, an intelligent regularizer should take into account the inherent similarity between these patches. To explain the difference between the TV and GTV priors we use a point model as shown in Fig. \ref{fig:demo}, where each point corresponds to a patch in the FBP. Since `a' and `b' are not spatially co-located, the total variation prior does not establish any connection between these patches. Thus, TV fails to exploit the similarity between these patches throughout the algorithm. This leads to slightly different textures for the two patches, as shown in the 3rd row of Fig. \ref{fig:reconstructions}. Now consider the case of GTV. \nauman{Even though the intial estimate of graph $\G$ is obtained from the noisy estimate of sample, \textit{i.e}, the FBP, patches `a' and `b' still possess enough structural resemblance to be connected together by an edge (even if it is weak) in the graph. Now, if the graph is kept fixed which is the case of GTV, one still obtains a better result as compared to CSTV, as shown in the 4th row of Fig. \ref{fig:reconstructions}. This is due to the fact that the important connections are established by the graph $\G$ and  similarity of patches is not restricted to spatially co-located patches only. This is also obvious from the intensity profile analysis in the 4th row of Fig. \ref{fig:reconstructions}. Finally, we discuss the case of AGTV, where the graph $\G$ is updated in every iteration of the algorithm. Obviously, every iteration of the algorithm leads to a cleaner sample and updating the graph $\G$ is only going to make the connection between the patches `a' and `b' stronger.  This leads to significantly better result than CSTV and GTV as shown in Fig. \ref{fig:reconstructions} and \ref{fig:reconstructions2}. Note that the patches `a' and `b' possess almost the same structure at the end of AGTV.}

\begin{figure*}
\centering
\includegraphics[width=\textwidth,height=13.5cm,keepaspectratio]{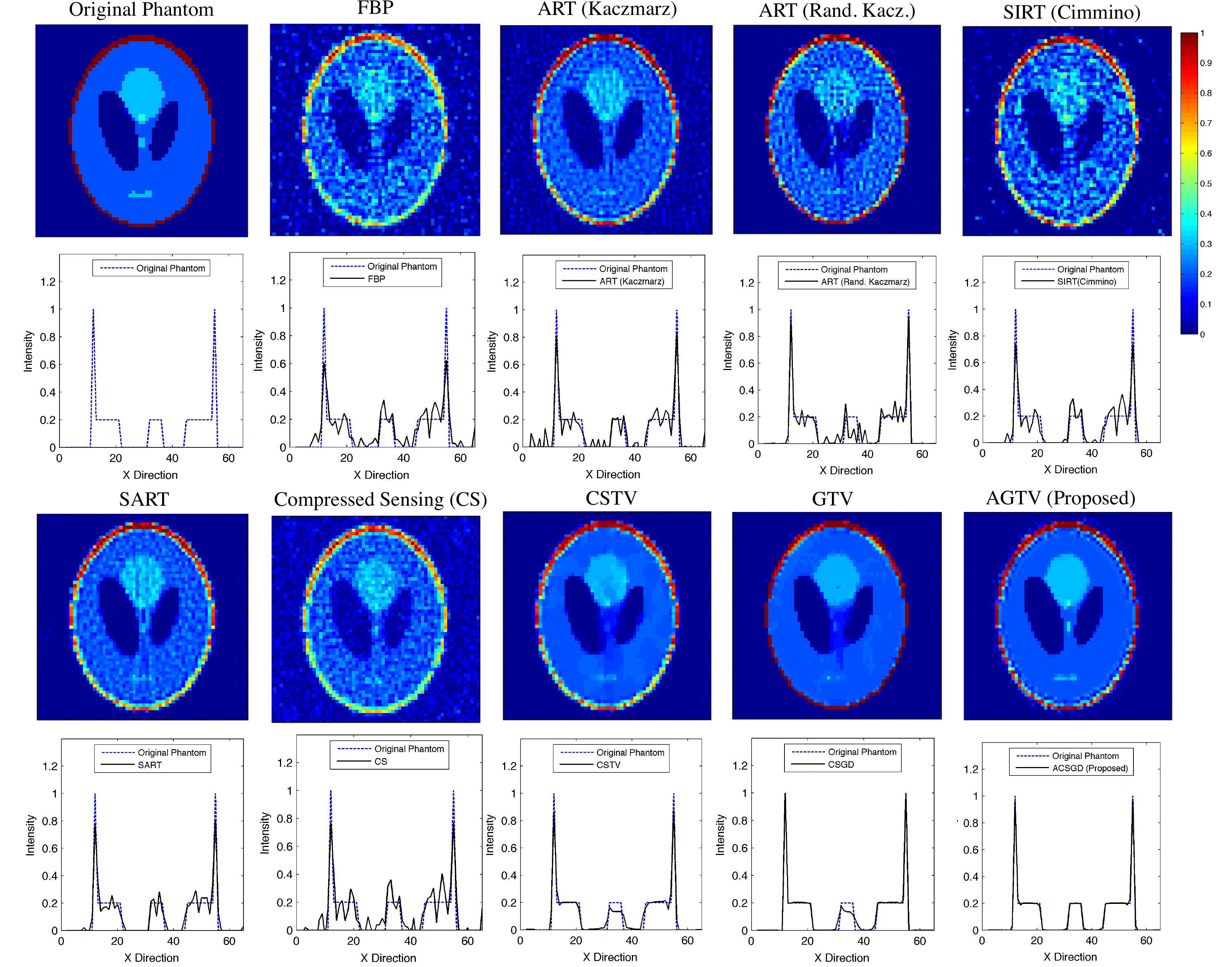}
\caption{Comparative analysis of reconstructing Shepp-Logan using various reconstruction methods. The sinogram of a $64\times64$ Shepp-Logan phantom corrupted with 10\% Poission noise was reconstructed using FBP (Linearly interpolated, Cropped Ram-Lak filter); ART (Kaczmarz/Randomized Kaczmarz, Relaxation Parameter $(\eta)$ = 0.25, Prior: FBP, Stopping Criteria = 100 iterations); SIRT (Cimmino/SART, $(\eta)$ = 0.25, Prior: FBP, Stopping Criteria = 100 iterations); CS (500 Iterations, Prior: FBP); CSTV ($\lambda=0.5$, $\gamma=0.1$, Prior: FBP, Stopping Criteria = 100 iterations); GTV ($\lambda=0.5$, $\gamma=0.2$, Prior: Patch Graph from FBP, Stopping Criteria = 100 iterations); AGTV ($\lambda=0.5$, $\gamma=1$, Prior: Patch Graph from FBP updated every iteration, $I$ and $J$ in Algorithm 1 set to 30). AGTV clearly gives a better intensity profile as compared to all other methods while preserving the edges.}
\label{fig:reconstructions}
\vspace{-1.2em}
\end{figure*}

\begin{figure*}
\centering
\includegraphics[width=\textwidth]{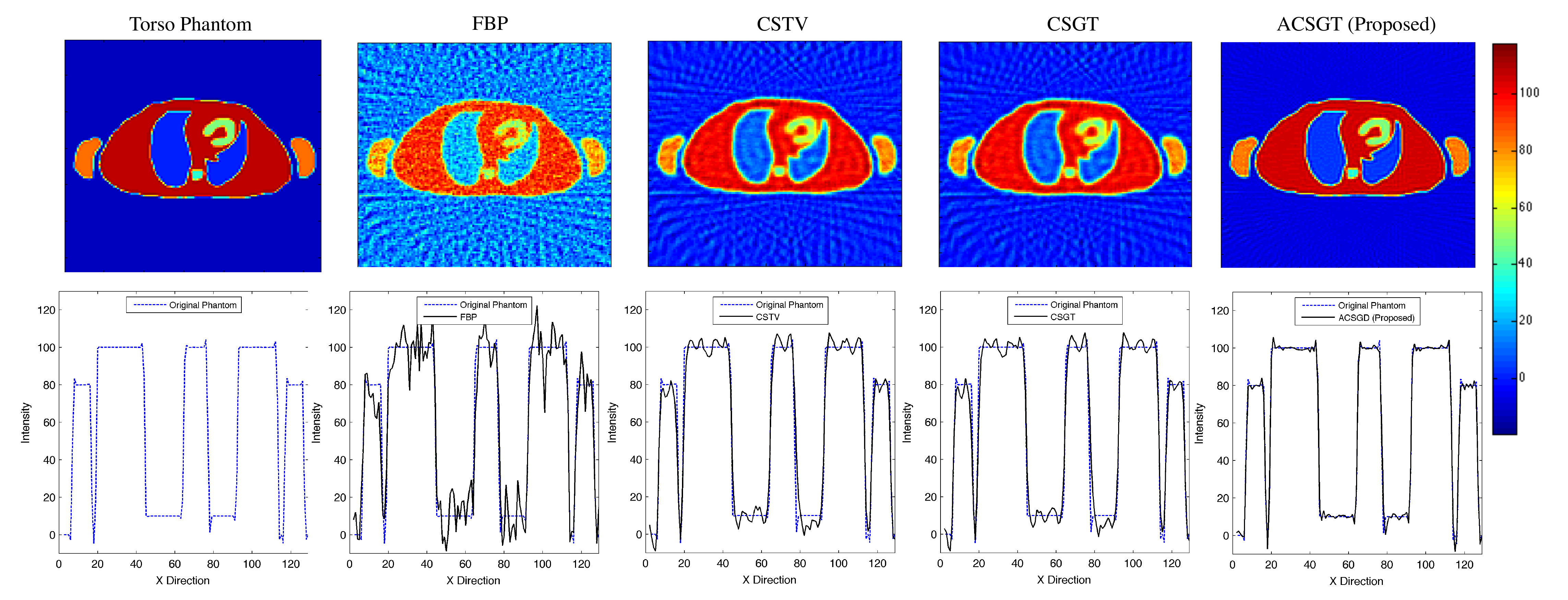}
\caption{Comparative analysis of reconstructing a Torso phantom using various reconstruction methods. The sinogram of a $128\times128$ Torso phantom corrupted with 5\% Gaussian Random noise was reconstructed using FBP (Linearly interpolated, Cropped Ram-Lak filter); CSTV ($\lambda=0.5$, $\gamma=0.1$, Prior: FBP, Stopping Criteria = 100 iterations); GTV ($\lambda=0.5$, $\gamma=0.2$, Prior: Patch Graph from FBP, Stopping Criteria = 100 iterations); AGTV ($\lambda=0.5$, $\gamma=0.1$, Prior: Patch Graph from FBP updated every iteration, $I$ and $J$ in Algorithm 1 set to 30).}
\label{fig:reconstructions2}
\vspace{-2em}
\end{figure*}


It is possible to appreciate this visually as the phantom obtained via AGTV is very similar to the original phantom. Furthermore, a comparison of the intensity profiles of the two phantoms also reveals the same fact. The next best result is obtained by CSGT. Algorithmically, the only difference between CSGT and AGTV is the regular graph update step in the latter, which tends to make the final reconstruction more faithful to the original phantom. CSTV also obtains a reasonable reconstruction, though worse than AGTV. CS alone however, has a poor performance. This is not surprising, as for the tomography applications,  CS has been mostly used in combination with TV, as it alone does not preserve the \fn{Gradient Magnitude Image (GMI)}. It is also interesting to note that the performance of AGTV saturates after 90 projections for each of the three cases, i.e, the reconstruction error does not improve if the number of projections are increased. Furthermore, for each of the three noise cases one can observe that the drop in the reconstruction error from 50 to 90 projections is not significant. Although, the same observation can be made about CSGT, the error is a always higher than AGTV. All the other methods, perform far worse than  AGTV. \fn{Moreover, a frequency analysis based on Fourier ring correlation using the Shepp-Logan phantom showed that AGTV preserved more higher frequency details as compared to other methods.}

\section*{B. Hyperparameter tuning}
{Our model has two hyper-parameters, $\lambda$ for tuning the sparsity of CS based reconstruction and $\gamma$ to tune the amount of smoothing and denoising in the reconstruction. While, these are model hyper-parameters and need tuning, the graph parameter $\K$, \textit{i.e}, the number of nearest neighbors is quite easy to set for our application. This is shown in Fig. \ref{fig:paramsK} where we perform a small experiment corresponding to the reconstruction of a $32 \times 32$ Shepp-Logan phantom from 36 projections $b \in \Rbb^{36}$ using the pre-tuned parameters $\lambda = 0.1, \gamma = 5$ for different values of $\K$ ranging from 5 to 50. \fn{The results clearly show that the reconstruction is quite robust to the choice of $\K$, with a small error variation.} Thus, $\K$ is easy to set for our application. As the complexity of our proposed algorithm scales with the number of edges $|\E|$ in the graph $\G$ and $|\E| \approx \K n^2$, it is recommended to set $\K$ as small as possible. \fn{However, a very small $\K$ might lead to many disconnected components in the graph $\G$. On the other hand, a very large $\K$ might increase the time required for the algorithm to converge and reduce the computational advantage we have over the NLTV method.} In order to show the variation of reconstruction error with $(\lambda, \gamma)$ grid, we perform another experiment for the reconstruction of the Shepp-Logan phantom of size $32 \times 32$ from 36 projections. For this experiment we keep $\K = 10$ and perform the reconstruction for every pair of parameter values in the tuple $(\lambda, \gamma)$, where $\lambda \in (0.1,1)$ and $\gamma \in (0.1,10)$. The reconstruction error grid is shown in Fig. \ref{fig:paramsK}. The minimum error $0.11$ occurs at $\lambda = 0.2, \gamma = 0.1$. It is also interesting to note that the error increases gradually with an increase in the parameter values. \fn{These representative hyperparameter tuning experiments are for demonstration and individual parameters were tuned for each experiment.}

 \begin{figure*}[h]
\centering
\includegraphics[width=5in]{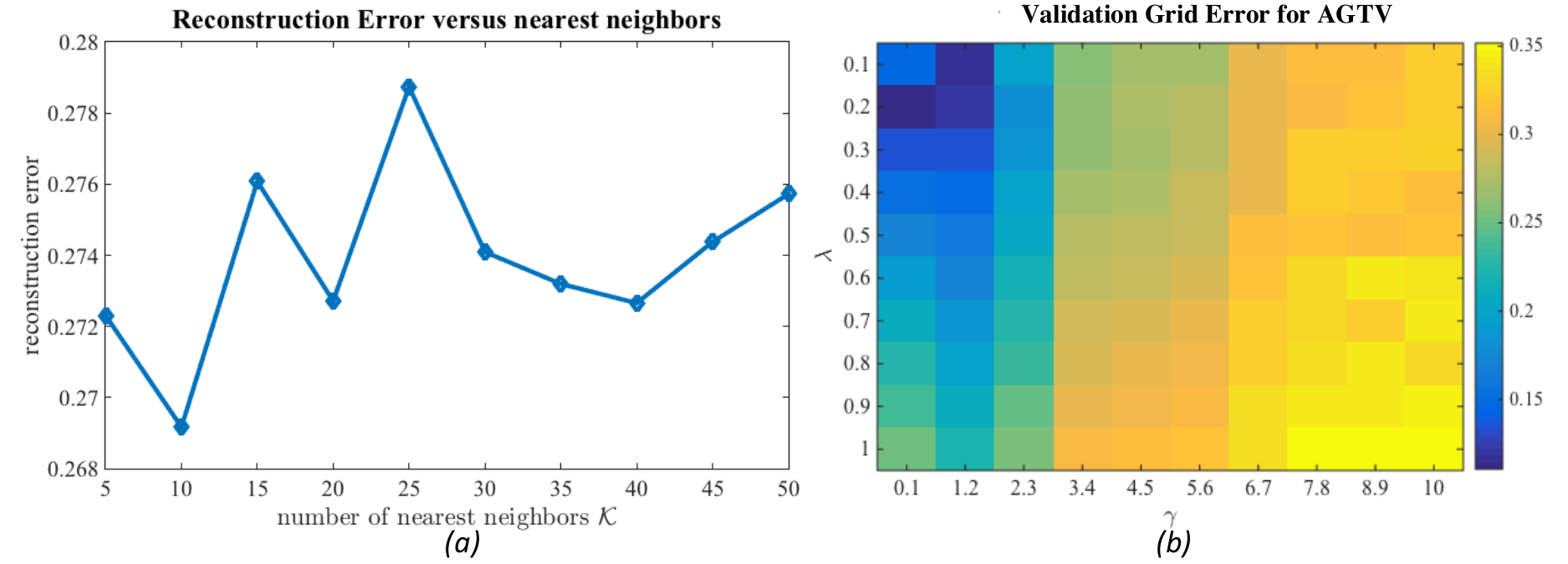}
\caption{a) A small experiment corresponding to the reconstruction of a $32 \times 32$ Shepp-Logan phantom from 36 projections $b \in \Rbb^{36}$ using the pre-tuned parameters $\lambda = 0.1, \gamma = 5$ for different values of $\K$ ranging from 5 to 50. The results clearly show that the reconstruction is quite robust to the choice of $\K$, with a small error variation. b) A small experiment corresponding to the reconstruction of a $32 \times 32$ Shepp-Logan phantom from 36 projections $b \in \Rbb^{36}$ using the full parameter gird $\lambda = (0.1, 1), \gamma = (0.1,10)$ for a fixed value of $\K = 10$. The minimum clustering error occurs at $\lambda = 0.2, \gamma = 0.1$. The results clearly show that the reconstruction error increases smoothly with the parameters. }
\label{fig:paramsK} 
\vspace{-1.7em}
\end{figure*}

\section*{C. Results with Real Data}

 \begin{figure*}[h]
\centering
\includegraphics[width=6.5in]{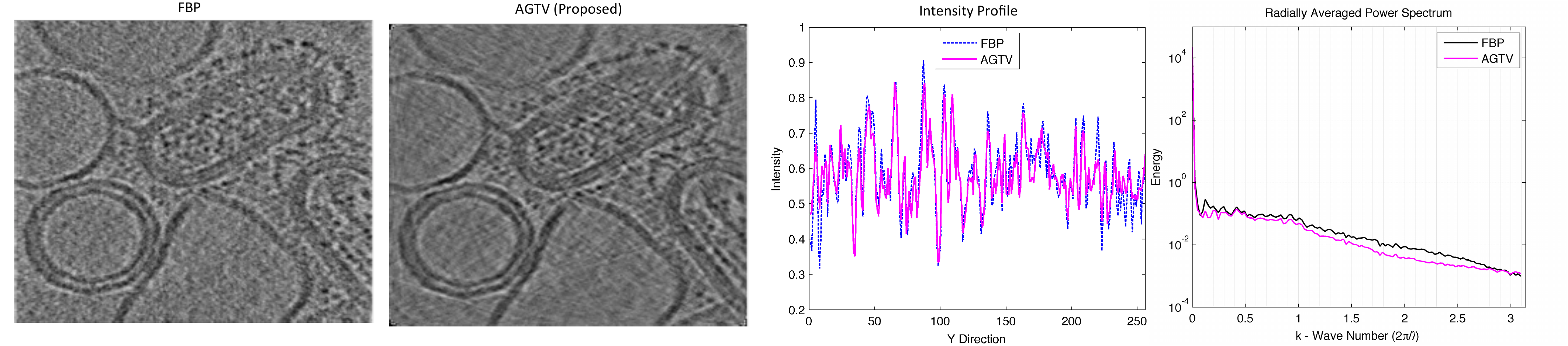}
\caption{\nauman{Cryo-ET data reconstructed using FBP (with sinogram denoising) and AGTV with their corrosponding Radially Averaged Power Spectrums (RAPS) and intensity profiles. Raw data used is from an Influenza virus available from open-source EM-Data-Base Entry: 4067. \fn{The reconstructed images have streaking artifacts because of missing data due to the missing wedge and limited projections.}}}
\label{fig:pan} 
\vspace{-1.7em}
\end{figure*}

\section*{D. Shortcomings \& Limitations}
The proposed AGTV method has proven to produce much better reconstructions as compared to the state-of-the-art CSTV method. Although, the proposed method is computationally far less cumbersome than NLTV, it still suffers from a few problems which we discuss in this section. The computational complexity of the proposed method is $\mathcal{O}(I(J \K n^2 + n^2\log(n^2)))$. The main computational burden is offered by the graph construction, which needs to be performed every $J$ iterations. Thus, the method still suffers from a high complexity because of the double loop and regular graph updates. The complexity of graph construction can be reduced by using a parallel implementation of FLANN. The degree of parallelism can be increased at the cost of increasing approximation in the estimation of nearest neighbors. As a result of this the graph $\G$ will be different every time the FLANN algorithm is run. However, this does not effect the quality of the graph and for tomographic applications, negligible loss in the performance was observed. It is obviously of interest to reduce the number of inner iterations $J$ and the complexity of the operations in the for loop. Tuning the hyperparameters is another short-coming of the proposed method.  

\end{document}